\documentclass[twocolumn]{article}
\usepackage[utf8]{inputenc}
\usepackage{graphicx}
\usepackage{amsmath}
\usepackage{url}
\usepackage{caption}
\usepackage{geometry}
\usepackage{times}

\makeatletter
\def\@maketitle{%
  \newpage
  \null
  \vskip 2em%
  \begin{center}%
  \let \footnote \thanks
    {\LARGE \@title \par}%
    \vskip 1.5em%
    {\large
      \lineskip .5em%
      \begin{tabular}[t]{c}%
        \@author
      \end{tabular}\par}%
    \vskip 1em%
  \end{center}%
  \par
  \vskip 1.5em}
\makeatother

\makeatletter
\renewenvironment{thebibliography}[1]{%
 \section*{\refname}%
 \@mkboth{\MakeUppercase\refname}{\MakeUppercase\refname}%
 \list{\@biblabel{\@arabic\c@enumiv}}%
      {\settowidth\labelwidth{\@biblabel{#1}}%
       \leftmargin\labelwidth
       \advance\leftmargin\labelsep
       \@openbib@code
       \usecounter{enumiv}%
       \let\p@enumiv\@empty
       \renewcommand\theenumiv{\@arabic\c@enumiv}}%
 \sloppy
 \clubpenalty4000
 \@clubpenalty \clubpenalty
 \widowpenalty4000%
 \sfcode`\.\@m}
 {\def\@noitemerr
   {\@latex@warning{Empty `thebibliography' environment}}%
  \endlist}
\makeatother

\title{Ghost Policies: A New Paradigm for Understanding and Learning from Failure in Deep Reinforcement Learning}

\author{Xabier Olaz$^{1}$\\
\small $^{1}$Department of Computer Science, Public University of Navarre, Pamplona, Spain}

\date{}

\begin{document}

\maketitle

\begin{abstract}
Deep Reinforcement Learning (DRL) agents often exhibit intricate failure modes that are difficult to understand, debug, and learn from. This opacity hinders their reliable deployment in real-world applications. To address this critical gap, we introduce ``Ghost Policies,'' a concept materialized through Arvolution, a novel Augmented Reality (AR) framework. Arvolution renders an agent's historical failed policy trajectories as semi-transparent ``ghosts'' that coexist spatially and temporally with the active agent, enabling an intuitive visualization of policy divergence. Arvolution uniquely integrates: (1) AR visualization of ghost policies, (2) a behavioural taxonomy of DRL maladaptation, (3) a protocol for systematic human disruption to scientifically study failure, and (4) a dual-learning loop where both humans and agents learn from these visualized failures. We propose a paradigm shift, transforming DRL agent failures from opaque, costly errors into invaluable, actionable learning resources, laying the groundwork for a new research field: ``Failure Visualization Learning.''
\end{abstract}

\section*{Introduction}

Deep Reinforcement Learning (DRL) has achieved unprecedented success in domains ranging from games\cite{silver2016mastering, mnih2013playing, vinyals2019grandmaster} to robotics\cite{kalashnikov2018qt, akkaya2019solving}. Despite these advances, the ``black-box'' nature of many DRL agents, especially when they fail, remains a significant obstacle. While the underlying theoretical causes of failure, such as distributional shift\cite{verma2024counterfactual, rabinowitz2019human} or reward hacking\cite{pan2022effects, skalse2022invariance}, are increasingly understood, a critical gap persists in our ability to observe and understand \textit{how} these failures manifest behaviourally during real-time adaptation.

Current tools for failure analysis offer limited support. Standard tools like TensorBoard provide abstract data plots, useful for monitoring high-level metrics but offering little intuition into the spatiotemporal behavioural execution of an agent's policy\cite{tensorflow2015-whitepaper}. This limitation underscores the need for more intuitive, information-rich tools to bridge the gap between abstract metrics and concrete agent behaviour.

\subsection*{State of the Art and Identified Research Gaps}
Our analysis reveals four critical gaps this work aims to address.

\textbf{GAP 1: The ``Ghost Policy'' concept does not exist.}
Existing work on counterfactual explanations focuses on discrete, alternative outcomes. For example, methods proposed by vanilla counterfactual explanation (VCE) or counterfactual outcome visualization (COViz) show what would have happened if an agent took a different action at a specific state\cite{olson2021counterfactual, amitai2023explaining}. These methods provide short-term, local alternatives, but not a visualization of a \textit{continuous, parallel historical policy} executing alongside a live agent. The concept of a ``Ghost Policy''---visualizing the entire trajectory of a past policy as it would unfold from a past state, coexisting with the live agent---is fundamentally novel. It shifts the question from ``What if?'' at a single point, to ``How does this agent's current long-term behaviour compare to its past self?''

\textbf{GAP 2: Behavioural taxonomies of DRL collapse do not exist.}
Current DRL failure taxonomies primarily catalogue software bugs or algorithmic errors. For instance, the taxonomy by Nikanjam et al. catalogues issues like ``RL-specific bugs such as lack of exploration'' or ``DNN-specific bugs such as vanishing gradient''\cite{nikanjam2021taxonomy}. Tools like RLExplorer diagnose these predefined faults by monitoring behavioural \textit{symptoms}\cite{nikanjam2024toward}. What is missing is a systematic classification of \textit{how} policies behaviourally maladapt, especially when facing novel environmental challenges. Emergent maladaptive behaviours (e.g., catatonic freezing, obsessive loops) are not formally categorized, yet they represent a higher level of abstraction than code-level faults.

\textbf{GAP 3: AR is not used as a scientific instrument for RL research.}
RL debugging tools are typically non-immersive\cite{tensorflow2015-whitepaper}. While AR/VR applications are emerging, they serve different purposes, such as robot data collection via teleoperation (e.g., ARMADA\cite{armada2024}) or industrial training simulation\cite{popov2023simulated}. No existing framework leverages AR as a \textit{scientific instrument} to immersively study an agent's policy adaptation and collapse in real-time. AR offers unique advantages for understanding complex spatiotemporal agent behaviour, including spatial co-presence and temporal superposition of multiple policies, which are difficult to achieve on 2D screens.

\textbf{GAP 4: Systematic human disruption is not a research method for cataloguing DRL failure modes.}
Current approaches to test RL agent robustness involve automated adversarial attacks, which apply algorithmic perturbations to observations or actions\cite{li2023adversarial, zhang2023adversarial}. Human-in-the-loop RL (HITL-RL), conversely, typically focuses on humans providing feedback to \textit{improve} agent training\cite{schuessler2024human, park2023improving}. The gap is the absence of a formal methodology for using \textit{intentional, creative, and systematic human-generated disruptions} as a scientific method to rigorously probe an agent's adaptive capabilities and catalogue its behavioural failure modes. Humans can design more nuanced and contextually relevant challenges than purely algorithmic perturbations, enabling the discovery of a richer set of failure modes.

\begin{table*}[t]
\centering
\caption{Comparison of Arvolution with Existing DRL Explanation and Debugging Approaches.}
\label{tab:comparison}
\scriptsize
\setlength{\tabcolsep}{3pt}
\begin{tabular}{l|p{1.8cm}|p{1.8cm}|p{1.8cm}|p{1.8cm}|p{1.8cm}|p{1.8cm}}
\hline
\textbf{Feature} & \textbf{Arvolution (Proposed)} & \textbf{COViz\cite{olson2021counterfactual, amitai2023explaining}} & \textbf{TensorBoard\cite{tensorflow2015-whitepaper}} & \textbf{RLExplorer\cite{nikanjam2024toward}} & \textbf{Adversarial Attacks\cite{li2023adversarial, zhang2023adversarial}} & \textbf{HITL (Training)\cite{schuessler2024human, park2023improving}} \\
\hline
\textbf{Visualization Modality} & Immersive AR, Spatio-temporal & 2D Visual (Side-by-side) & 2D Graphs, Non-immersive & Log-based, Textual warnings & N/A & Varies (GUI/Text) \\
\hline
\textbf{Policy Visualization} & Continuous, Parallel (Live + Ghosts) & Discrete Alternative Trajectories & Aggregated Statistics & Training Dynamics Metrics & N/A & Live Agent Behaviour \\
\hline
\textbf{Temporal Scope} & Historical \& Recent Policies & Short-term ($k$ steps) & Training Epochs & Training Traces & N/A & Current Interaction \\
\hline
\textbf{Failure Focus} & Behavioural Mal-adaptation (How) & Local Decision Consequences & Performance Metrics & Software Fault Symptoms & Induce Failure (Robustness) & Guide to Success \\
\hline
\textbf{Interaction Method} & Systematic Human Disruption & N/A & Data Exploration & Automated Diagnosis & Automated Perturbation & Human Feedback/Demo \\
\hline
\textbf{Learning from Failure} & Dual-Loop (Human \& Agent learn from visualized ghosts) & Primarily Human Understanding & Human Analysis & Human Debugging Guidance & Agent Robustness Training & Agent Learns from Guidance \\
\hline
\end{tabular}
\end{table*}

\section*{Results}

\subsection*{The Arvolution Framework}
Arvolution is a multi-dimensional system designed to make DRL failure transparent and actionable. It integrates four core components.

\subsubsection*{Multi-Layer Temporal Visualization in AR}
The user observes multiple policy layers simultaneously:
\begin{itemize}
    \item \textbf{Layer 1: Current Agent.} The live, fully opaque agent.
    \item \textbf{Layer 2: Recent Ghost.} A semi-transparent policy from a recent episode.
    \item \textbf{Layer 3: Historical Ghost.} A highly transparent policy from a distant episode.
    \item \textbf{Layer 4: Pre-disruption Ghost.} A distinctly coloured ghost showing the optimal policy before a failure-inducing disruption.
\end{itemize}

\begin{figure*}
  \centering
  \includegraphics[width=0.8\textwidth]{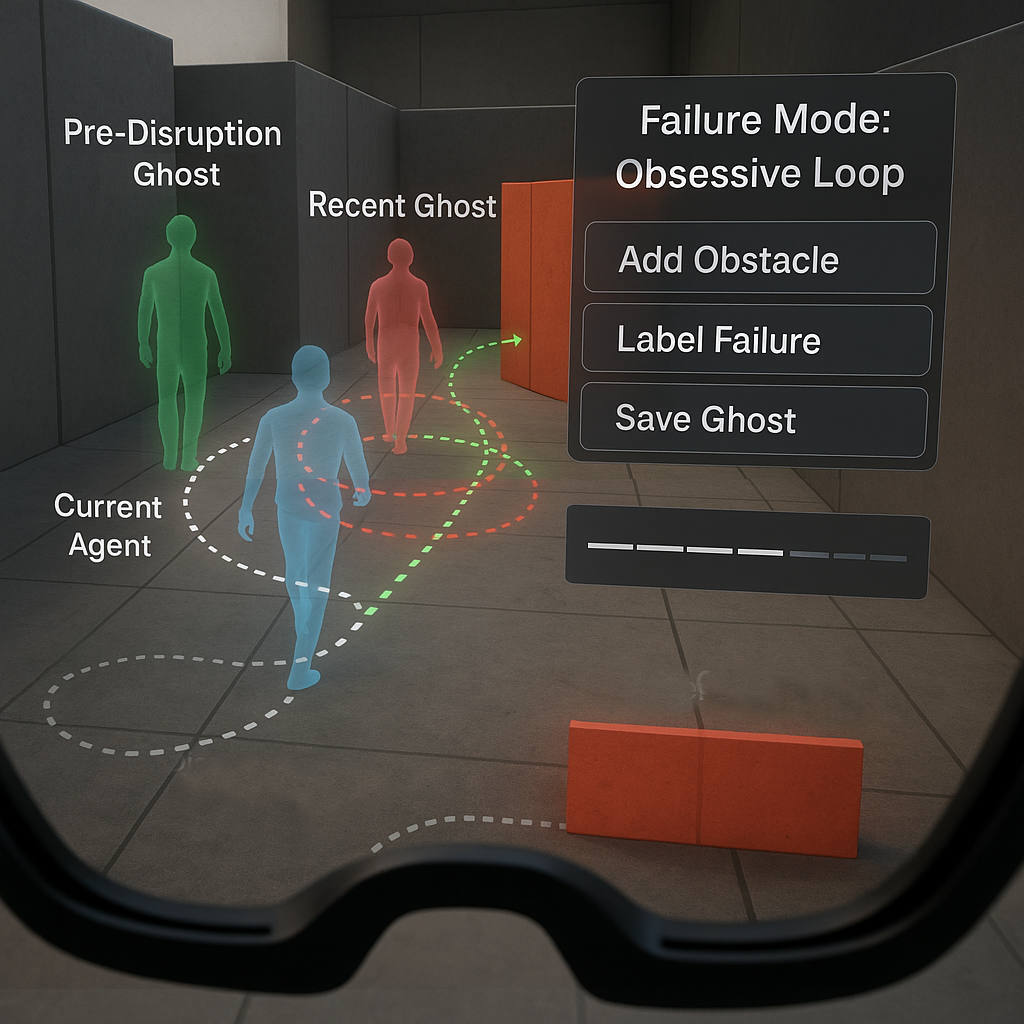}
  \caption{\textbf{Conceptual visualization of the Arvolution Framework.} The user, wearing an AR headset, observes a live agent (blue). They simultaneously see a ``Pre-disruption Ghost'' (green) showing the previously successful path, and a ``Recent Ghost'' (red) demonstrating the current, failing policy trajectory characterized by an obsessive loop. The AR interface allows the human to introduce disruptions and label the observed failure mode.}
  \label{fig:ar_layers}
\end{figure*}

\subsubsection*{Behavioural Taxonomy of Maladaptation}
We propose the first behavioural taxonomy focused on how DRL agents fail during adaptation. Initial categories include:
\begin{itemize}
    \item \textbf{Catatonic Collapse:} The agent ceases meaningful action.
    \item \textbf{Manic Oscillation:} The agent rapidly alternates between contradictory actions.
    \item \textbf{Obsessive Loop:} The agent becomes trapped in a suboptimal policy loop.
    \item \textbf{Gradual Drift:} A slow, incremental deviation from an optimal policy.
    \item \textbf{Policy Fragmentation:} The agent's behaviour becomes disjointed and incoherent.
\end{itemize}

\subsubsection*{Human Disruption Protocol}
To systematically study these failure modes, we developed a standardized protocol for human-in-the-loop disruption. Proposed disruption types include `obstacle\_placement', `goal\_relocation', `physics\_alteration', `reward\_inversion', and `sensory\_occlusion'.

\subsubsection*{The Dual-Learning Loop}
Arvolution's core contribution is a dual-learning loop for both human and agent learning. For the agent, we propose an active learning mechanism where the policy, $\pi$, is conditioned on the current state, $s_t$, and a representation of relevant historical failures, retrieved from a database of stored ghosts, $D$.

\begin{center}
\fbox{\parbox{0.9\columnwidth}{
\footnotesize
\texttt{\# Conceptual Pseudocode}\\
\texttt{def get\_action(current\_state):}\\
\texttt{\ \ \ \ \# Retrieve ghosts of past failures}\\
\texttt{\ \ \ \ \# in similar states}\\
\texttt{\ \ \ \ relevant\_ghosts = retrieve\_ghosts(}\\
\texttt{\ \ \ \ \ \ \ \ current\_state, D)}\\
\texttt{\ \ \ \ \# Extract actions that led to failure}\\
\texttt{\ \ \ \ failure\_actions = get\_failure\_actions(}\\
\texttt{\ \ \ \ \ \ \ \ relevant\_ghosts)}\\
\texttt{\ \ \ \ \# Condition policy to avoid}\\
\texttt{\ \ \ \ \# failure actions}\\
\texttt{\ \ \ \ action = policy(current\_state,}\\
\texttt{\ \ \ \ \ \ \ \ avoid=failure\_actions)}\\
\texttt{\ \ \ \ return action}
}}
\end{center}

\section*{Discussion}

The Arvolution framework represents a paradigm shift, re-framing agent failures as a valuable data resource. It moves DRL analysis from post-mortem debugging to real-time, intuitive understanding, making the agent's adaptation process tangible. This establishes a new field we term ``Failure Visualization Learning.''

Our research plan is twofold. Phase 1 will focus on a Minimum Viable Product (MVP) to validate the qualitative insight provided by Ghost Policies. This will mitigate key technical risks and produce an initial dataset of labeled failure trajectories. Phase 2 will scale the MVP into a full-featured, open-source framework and implement the dual-learning loop. The primary goal of Phase 2 is to quantitatively demonstrate that agents can improve their learning speed and final performance by actively learning from their past visualized failures. Positive quantitative results would validate the ``failures as a resource'' concept and establish Arvolution as a significant contribution to DRL methodology.

\section*{Methods}

\subsection*{DRL Environment and AR Implementation}
Experiments will be conducted within the Unity Engine, using the ML-Agents Toolkit\cite{juliani2018unity} and a PyTorch backend. The AR visualization will be developed for a standalone headset (e.g., Meta Quest 3) using AR Foundation. Trajectory data will be streamed in real-time to the AR application.

\subsection*{Taxonomy and Disruption Protocol}
The behavioural taxonomy will be developed iteratively. In Phase 1, researchers will use the Arvolution MVP to manually label failure trajectories induced by the disruption protocol, calculating inter-rater reliability to ensure consistency. In Phase 2, a machine learning classifier will be trained on this labeled dataset to automate failure classification.

\subsection*{Dual-Learning Loop Evaluation}
In Phase 2, we will quantitatively evaluate the dual-learning loop. The performance of agents trained with Arvolution will be compared against standard baselines (e.g., the same RL algorithm without the failure-learning mechanism). Key metrics will include sample efficiency (e.g., reduction in training episodes to reach a performance criterion), final asymptotic performance, and robustness to novel disruptions.

\end{document}